%% file: main.tex
\documentclass[letterpaper, 10 pt, conference]{ieeeconf}
\IEEEoverridecommandlockouts

\title{\LARGE \bf
WMNav: Integrating Vision-Language Models into World Models for Object Goal Navigation
}

\author{
    Dujun Nie\textsuperscript{1,$^*$}\thanks{$^*$This author contributed equally to this work.},
    Xianda Guo\textsuperscript{2,$^*$,$^{\dagger}$}\thanks{$^\dagger$Project Leader},
    Yiqun Duan\textsuperscript{3},
    Ruijun Zhang\textsuperscript{1},
    Long Chen\textsuperscript{1,4,5,$^\ddagger$} \thanks{$^\ddagger$Corresponding Author}\\
    \textsuperscript{1} Institute of Automation, Chinese Academy of Sciences \textsuperscript{2} School of Computer Science, Wuhan University \\ \textsuperscript{3} School of Computer Science, University of Technology Sydney
    \textsuperscript{4} IAIR, Xi'an Jiaotong University
    \textsuperscript{5} Waytous \\
\texttt{\{niedujun2024,zhangruijun2023,long.chen\}@ia.ac.cn; }\\
\texttt{xianda\_guo@163.com; duanyiquncc@gmail.com} 
\vspace{-5mm}
}

\usepackage{caption} 
\usepackage{graphicx}
\usepackage{booktabs} 
\usepackage{amsmath}
\usepackage{amssymb}
\usepackage{algorithm}
\usepackage{algorithmic}
\usepackage{paralist}
\usepackage{multirow}
\usepackage{dashrule}
\usepackage{pifont}
\usepackage{textcomp}
\usepackage[T1]{fontenc}
\usepackage{hyperref}
\usepackage{cleveref}
\usepackage[table]{xcolor} 
\usepackage{xcolor}
\usepackage{color}
\definecolor{lightgray}{gray}{.9}
\usepackage{array}
\newcolumntype{I}{!{\vrule width 1pt}}
\makeatletter
\newcommand{\thickhline}{%
    \noalign {\ifnum 0=`}\fi \hrule height 1pt
    \futurelet \reserved@a \@xhline
}
\makeatother

\begin{document}
\maketitle

\thispagestyle{empty}
\pagestyle{empty}

\input{seg/0_abstract}

\input{seg/1_intro.tex}

\input{seg/2_related.tex}

\input{seg/3_method.tex}

\input{seg/4_experiments.tex}

\input{seg/5_conclusions.tex}

\input{seg/8_acknowledgment}

\bibliographystyle{IEEEtran}
\bibliography{main}

\end{document}

%% file: seg/0_abstract.tex
\begin{abstract}
Object Goal Navigation—-requiring an agent to locate a specific object in an unseen environment—-remains a core challenge in embodied AI.
Although recent progress in Vision-Language Model (VLM)-–based agents has demonstrated promising perception and decision-making abilities through prompting, none has yet established a fully modular world model design that reduces risky and costly interactions with the environment by predicting the future state of the world. We introduce WMNav, a novel World Model-based Navigation framework powered by Vision-Language Models (VLMs). It predicts possible outcomes of decisions and builds memories to provide feedback to the policy module. To retain the predicted state of the environment, WMNav proposes the online maintained Curiosity Value Map as part of the world model memory to provide dynamic configuration for navigation policy. By decomposing according to a human-like thinking process, WMNav effectively alleviates the impact of model hallucination by making decisions based on the feedback difference between the world model plan and observation. 
To further boost efficiency, we implement a two-stage action proposer strategy: broad exploration followed by precise localization. Extensive evaluation on HM3D and MP3D validates WMNav surpasses existing zero-shot benchmarks in both success rate and exploration efficiency (absolute improvement: +3.2\% SR and +3.2\% SPL on HM3D, +13.5\% SR and +1.1\% SPL on MP3D). Project
page: \href{https://b0b8k1ng.github.io/WMNav/}{https://b0b8k1ng.github.io/WMNav/}.
\end{abstract}

%% file: seg/1_intro.tex
\section{Introduction}
Effective navigation is a fundamental requirement for domestic robots, allowing them to access specific locations and execute assigned operations~\cite{objnav}. Zero-Shot Object Navigation (ZSON) is a critical component of this functionality, which demands that an agent locate and approach a target object from an unseen category through environmental understanding. This navigation capability proves indispensable for performing varied and sophisticated functions in practical home environments. The primary difficulty in ZSON stems from the need to employ broad semantic knowledge to direct movement with optimal efficiency while precisely identifying previously unencountered target objects.
 \begin{figure}[tbp]
  \centering
  \includegraphics[width=0.5\textwidth]{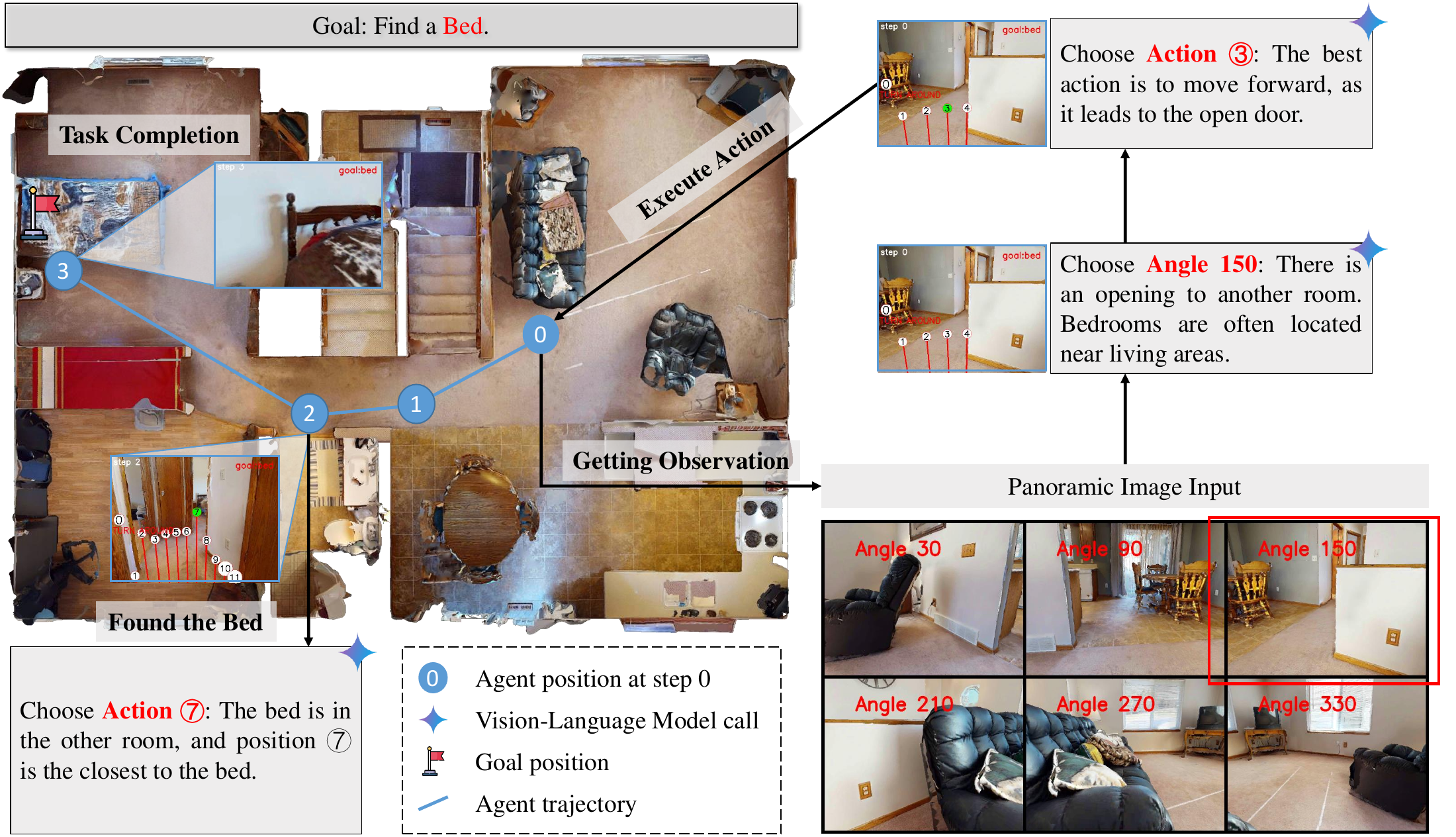}
     \vspace{-12pt}
   \caption{\textbf{World Model Navigation with VLM.} In object navigation, our model first estimates the goal's presence likelihood in each scene of the panoramic image (e.g., a bed typically resides in a living room, which is likely situated at the corridor's terminus), then plans intermediate subtasks and chooses the most appropriate action to execute.}
   \label{fig:overview}
    \vspace{-16pt}
\end{figure}

With the recent advancements in visual foundation models, the effectiveness of visual perception has experienced a substantial enhancement, achieving remarkable performance in zero-shot scene understanding. Existing navigation methods can be divided into two categories: network-based navigation~\cite{majumdar2022zson, chen2023zeroshot, gadre2023cows} and map-based navigation\cite{zhou2023esc, chen2023train, wu2024voronav, kuang2024openfmnav, zhong2024topv}. Network-based methods usually use reinforcement learning or imitation learning, which rely on trainable modules that need large amounts of data to train(such as policy learning)\cite{ovrl, ovrl2, pixnav}, resulting in significant computational resources and time for learning. Map-based methods construct maps to preserve semantic information about the scene and generate frontiers or waypoints for path planning. Their path planning heavily relies on accurate maps, which involve complex and time-consuming map construction processes. These methods fail to fully leverage the capabilities of VLMs trained with egocentric RGB images. Additionally, the selection strategies employed are often inflexible, frequently falling short of identifying optimal positions for decision-making. To address the limitations inherent in map-based planning, VLMNav\cite{vlmnav} introduces an action space generation method that leverages navigable regions in egocentric images, providing a more flexible decision space and fully exploiting the capabilities of VLMs. However, due to the limited field of view of egocentric images, capturing environmental information outside the immediate perspective remains a significant challenge. 

What's more, most existing methods, whether network-based methods or map-based methods, require actual interaction with the environment to achieve accurate scene understanding. These approaches can not leverage predictive information about future states and the outcomes of potential actions, limiting their ability to perform anticipatory planning and reasoning effectively in uncertain scenarios. VLFM\cite{vlfm} constructs a value map of the likelihood of each region to lead toward the out-of-view target object. It makes use of prediction information to some extent. Still, it uses BLIP-2\cite{blip2}, which pays more attention to the relevance of image-text pairs and has limited interaction and reasoning capabilities, which makes it difficult to cope with complex planning tasks. When navigating to a specific object, humans can effectively imagine the room layout and anticipate action outcomes based on visual cues and common design patterns (such as walking along the corridor often leads to a new room and going to the bathroom to find a sofa is unwise).

The world model is a computational representation of environment dynamics that equips agents with the ability to simulate interactions. By simulating action choices within the virtual environment, agents can explore potential outcomes safely without directly interacting with the environment. This method not only reduces collision with obstacles but also enhances the agent's efficiency in planning and exploring. However, the true challenge lies in creating a versatile world model that can faithfully capture the landscape of an indoor environment. ATD~\cite{atd} uses a dual-branch self-guided imagination policy based on large language models with a human-like left-right brain architecture. WebDreamer~\cite{webdreamer} uses LLMs as world models in web environments for model-based planning of web agents, demonstrating its effectiveness over other baselines. Can VLMs also function as world models? Given that VLMs are trained on vast amounts of first-person human perspective images, we hypothesize that they have acquired sufficient common sense knowledge to simulate the outcomes resulting from human choices and actions. With their extensive pre-trained knowledge spanning indoor layout and general reasoning ability, VLM has great potential to take on this task. 

Building on the key insight that VLMs inherently encode comprehensive knowledge about indoor layout and spatial relationships of objects, we propose WMNav as shown in \Cref{fig:overview}, which is a pioneering approach leveraging VLMs in a world model paradigm to enable efficient navigation in complex indoor environments. It receives panoramic scene information and uses VLMs to quantitatively predict future outcomes and finish all the perception, planning, reasoning, and controlling processes without task-specific training, pre-built maps, or prior knowledge of the surroundings(using detection or segmentation results).

Our contributions can be summarized as follows:
\begin{itemize}
\item We introduce a new direction for object goal navigation in a complex, unknown environment using a world model consisting of VLMs and novel modules.
\item We design an innovative memory strategy of predicted environmental states that employs an online Curiosity Value Map to quantitatively store the likelihood of the target's presence in various scenarios predicted by the world model. 
\item We propose a subtask decomposition with feedback and a two-stage action proposer strategy to enhance the reliability of VLM reasoning outcomes and improve exploration efficiency.
\item We achieve state-of-the-art results on the Zero-shot Object Navigation task and outperform benchmark methods on HM3D~\cite{ramakrishnan2021habitat} and MP3D~\cite{Matterport3D}.
\end{itemize}

%% file: seg/2_related.tex
\section{Related Work}\label{sec:related}
\subsection{Zero shot Object Goal Navigation}
Existing object navigation methods fall into two paradigms: supervised methods and zero-shot methods. Supervised approaches either train visual encoders with RL/IL policies~\cite{Treasure_Hunt,khandelwal2022:embodied-clip, Habitat-Web,chen2022learning} or build semantic maps from training data~\cite{min2022film,zheng2022jarvis}. While effective in known environments, these methods struggle with unseen objects/rooms due to training dependency. Recent zero-shot works address this via open-vocabulary scene understanding. Image-based methods map target objects to visual embeddings~\cite{majumdar2022zson,al2022zero,gadre2023cows}, while map-based approaches use frontier-based map~\cite{chen2023train,zhou2023esc,yu2023l3mvn,shah2023lfg, kuang2024openfmnav, zhong2024topv} or waypoint-based map~\cite{wu2024voronav} with LLM/VLM reasoning. Here, we focus on the zero-shot ObjectNav task in unknown environments using a novel world model framework.

\subsection{Foundation Model Guided Navigation}
Recent advances in embodied navigation highlight the complementary roles of Vision-Language Models (VLMs) and Large Language Models (LLMs), driven by their distinct strengths in visual grounding and common sense reasoning. LLMs enable zero-shot commonsense reasoning through object-room correlation prediction \cite{zhou2023esc}, semantic mapping \cite{yu2023l3mvn, wu2024voronav}, and chain-of-thought planning \cite{shah2023lfg, cai2023bridging}. However, LLMs are unable to process rich visual information and do spatial reasoning. VLMs like CLIP \cite{khandelwal2022:embodied-clip, majumdar2022zson, glip} address this by directly aligning visual observations with textual goals through multimodal embeddings, offering seamless integration with exploration policies through unified visual features. Compared to LLM-based methods, VLM-based methods~\cite{zhong2024topv, vlmnav} may be more efficient and elegant for target-driven navigation by avoiding language-mediated grounding gaps. This is due to their multimodal nature and better spatial reasoning capabilities~\cite{surds}. Our framework uses pure VLM, and unlike other approaches, we use the VLM to predict the state of the environment caused by actions.

\begin{figure*}[tbp]
  \centering
  \includegraphics[width=0.9\textwidth]{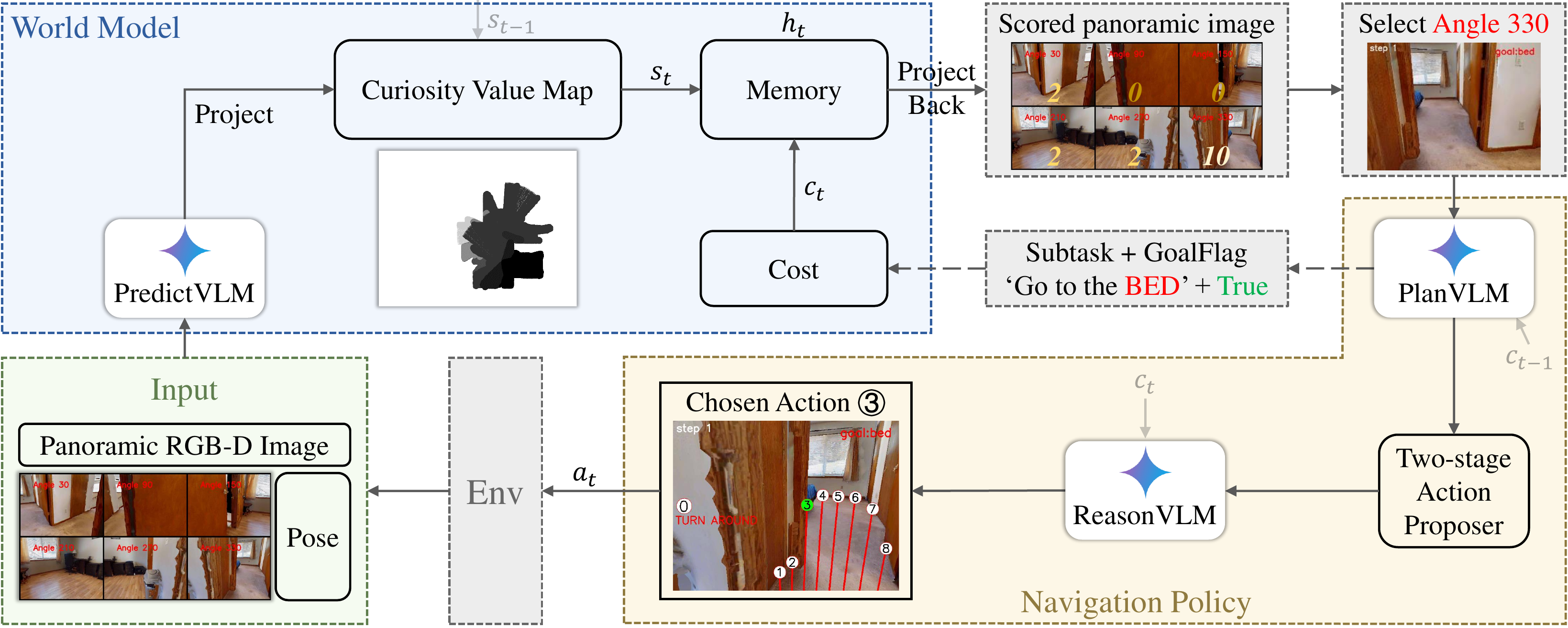}
   \caption{\textbf{The WMNav framework.} After acquiring the RGB-D panoramic image and pose information at step $t$, the PredictVLM first predicts the state of the world, and the state is merged with the curiosity value map $s_{t-1}$ from the previous step to get the current curiosity value map $s_t$. After that, the updated map projects the scores of each direction back onto the panoramic image, and the direction with the highest score is selected. Secondly, given the selected direction image, the new subtask and the goal flag are determined by PlanVLM and are stored in memory as cost $c_t$, and the memory $h_t$ is combined by $s_t$ and $c_t$. Finally, the two-stage action proposer annotates the action sequence on the selected image and sends it into ReasonVLM to obtain the final polar coordinate vector action $a_t$ for execution. Note that PlanVLM and ReasonVLM are configured by the cost $c_{t-1}$.}
   \label{fig:method}
     \vspace{-10pt}
\end{figure*}
\subsection{World Models}
World models originated in classical reinforcement learning with Dyna's simulated experiences \cite{sutton1991dyna}, evolving into frameworks for predicting state transitions to improve sample efficiency \cite{ha2018world, moerland2023model}. These models enable policy training in diverse domains and emphasize high-fidelity environment dynamics.
Recent work explores large language models (LLMs) as abstract world models, prioritizing task abstraction over precise simulation. Studies like \cite{kim2024cognitive} demonstrate LLMs' ability to leverage commonsense knowledge for planning in simple environments, while \cite{chae2024web} applies LLMs to web navigation. However, LLMs struggle with visually grounded decision-making. Therefore, it is more advantageous to integrate VLMs into world models in visual navigation tasks. VLMs combine visual grounding with semantic reasoning, enabling multimodal state understanding and adaptation to unseen environments through joint visual-textual prompts. Our approach takes VLM in a world model framework and involves it in all the processes of navigation, including predicting, planning, reasoning, action, etc.

%% file: seg/3_method.tex
\section{WMNav Approach} \label{sec:method}
\subsection{Task Definition} \label{sec:task def}
The traditional object goal navigation task requires the agent to explore an unknown indoor environment and navigate to an arbitrary instance $i$ within a given category $c$ (e.g., bed, sofa, toilet). The agent starts from a designated initial position. At each time step $t$, the agent takes an RGB-D observation $O_t$ of the surroundings and its real-time pose $P_t$. Then, the agent determines the action $a_t$ to find the goal. The task is deemed successful if the agent stops at a position within a predefined distance threshold $d_{thres}$ from the goal. While most previous works utilize a discrete action space, such as \{Stop, Move Forward, Turn Left, Turn Right, Look Up, Look Down\}, we adopt polar coordinates $(r_t, \theta_t)$ to represent the action $a_t$, where $\theta_t$ denotes the direction of the action and $r_t$ indicates the distance traveled by the action. 
\subsection{Overview}
  The framework of WMNav is depicted in~\Cref{fig:method}. A panoramic understanding is essential to make a comprehensive perception. To this end, the agent undergoes a series of rotations $\mathcal{A}=\{30^{\circ}, 90^{\circ}, 150^{\circ}, 210^{\circ}, 270^{\circ}, 330^{\circ}\}$ and captures six distinct RGB-D images which are converted into a panoramic image $I^{pan}_t$. In our framework, the world model consists of PredictVLM and the memory constructed by curiosity value map and cost. The world model does not receive any actual reward signals from the environment, which means it is only used to predict and simplify the future state of the environment. PredictVLM quantitatively predicts the likelihood of the goal's presence in each direction and projects the scores from a panoramic image to a top-down map. The map is then merged with the curiosity value map from the previous step and is stored in the memory. After that, the curiosity scores are projected back onto the panoramic image. Then, the direction in the panoramic image with the highest score is selected and sent to the navigation policy module. The navigation policy module has access to the reward information from the environment. For PlanVLM and ReasonVLM in the policy module, the cost (the previous step's subtask and the goal flag) is used to configure their prompts, thus optimizing the action output of the entire policy module without any fine-tuning of the VLMs.

\subsection{World Model} \label{sec:wmnav}
\begin{figure*}[t]
  \centering
  \includegraphics[width=\textwidth]{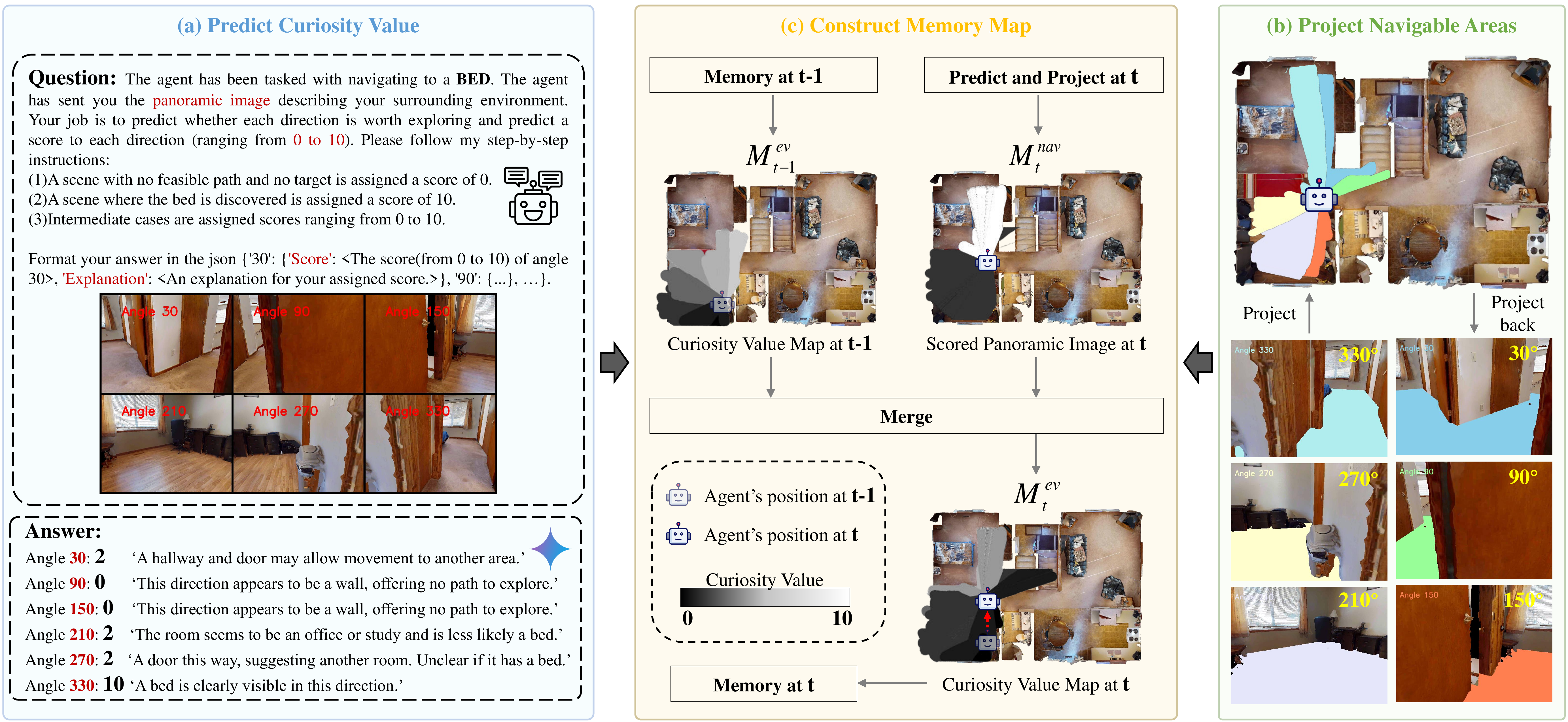}
    \vspace{-3ex}
   \caption{\textbf{Predict the Likelihood.} (a) The world model predicts the Curiosity Value for each direction in the panoramic image based on the likelihood of the goal's presence. (b) The mutual projection of the navigable area between the ego-centric and top-down view perspectives. (c) Curiosity Value Map construction: The predicted scores from the world model are projected onto the top-down map and then fused with the previous step's Curiosity Value Map.}
   \label{fig:Predict}
   \vspace{-13pt}
\end{figure*}
\subsubsection{VLM-based State Prediction} \label{sec:predictvlm}
The core capability of the world model is to estimate the state of the world, which is not provided by perception, and to predict possible future state changes. We employ a VLM as the predictor in the world model. To guide the VLM to make reasonable predictions about the indoor scene, we design a novel prompting strategy as illustrated in~\Cref{fig:Predict} (a). We use a panoramic image as the image prompt. PredictVLM is tasked with predicting each viewpoint's curiosity value, which represents the likelihood of the target's presence in each direction and scores from 0 to 10. The VLM outputs scores for each direction denoted as $Score_t(\alpha),(\alpha \in \mathcal{A})$. At each time step $t$, the panoramic image $I^{pan}_t$ is input to PredictVLM, which outputs scores $Score_t$ for each direction in the current panoramic view:
\begin{equation}
    Score_t=PredictVLM(I^{pan}_t)
\end{equation}
These scores can then be leveraged to construct the curiosity value map. 

\subsubsection{Curiosity Value Map Construction} \label{sec:cvm}
The curiosity value map $M^{cv}$ has a size of $map\_size\times map\_size \times 1$, where each pixel value represents the curiosity value of the position in the entire scene, ranging from 0 to 10. For regions that have been visited and found to be devoid of the goal object, the curiosity value is set to 0. For example, an observed bedroom that does not contain the goal television would have a curiosity value of 0. Regions where the goal can be directly discovered have a curiosity value of 10. For areas that have not yet revealed the goal but offer potential pathways to the goal, the curiosity value is set between 0 and 10 based on the VLM's imagination. 

The construction and update process is illustrated in~\Cref{fig:Predict} (c). Initially, all pixels in $M^{cv}_0$ are set to 10, indicating it is possible to find the target in all regions as the agent does not have any information about the entire scene. 
To convert the scores from the ego-centric perspective to the top-down view of the curiosity value map, we perform a projection transformation using depth information $D_t$ and pose $P_t$, similar to~\cite{vlmnav}~\cite{moma-kitchen}. We project the navigable area with scores $Score_t$ from the ego-centric view onto a top-down map $M^{nav}_t$ as in~\Cref{fig:Predict} (b):
\begin{equation}
    M^{nav}_t = Projection(Score_t)
\end{equation}
where $M^{nav}_t$ has the same size as $M^{cv}$. Then $M^{cv}_t$ ($s_t$ in~\Cref{fig:method}) is updated by combining $M^{nav}_t$ with the curiosity value map in the previous step $M^{cv}_{t-1}$ ($s_{t-1}$ in~\Cref{fig:method}):
\begin{equation}
    M^{cv}_{t}(u,v) = \min(M^{cv}_{t-1}(u,v), M^{nav}_{t}(u,v))
\end{equation}

\subsubsection{Cost} \label{sec:cost}
The cost module in the world model is used to provide environmental rewards. In our work, we use the subtask and the goal flag as the cost. For the detailed generation process of subtasks, refer to the Subtask Decomposition section (see Section~\ref{sec:sd}). The goal flag indicates whether the PlanVLM finds the goal in the selected image and is set to either True or False. The cost is fed into PlanVLM and ReasonVLM as part of their prompts to implicitly optimize the outputs in the navigation policy. PlanVLM receives the subtask from the previous step, and ReasonVLM receives the current subtask and the goal flag to switch between two configuration states.

\subsection{Subtask Decomposition} \label{sec:sd}
\begin{figure}[t]
  \centering
  \includegraphics[width=0.49\textwidth]{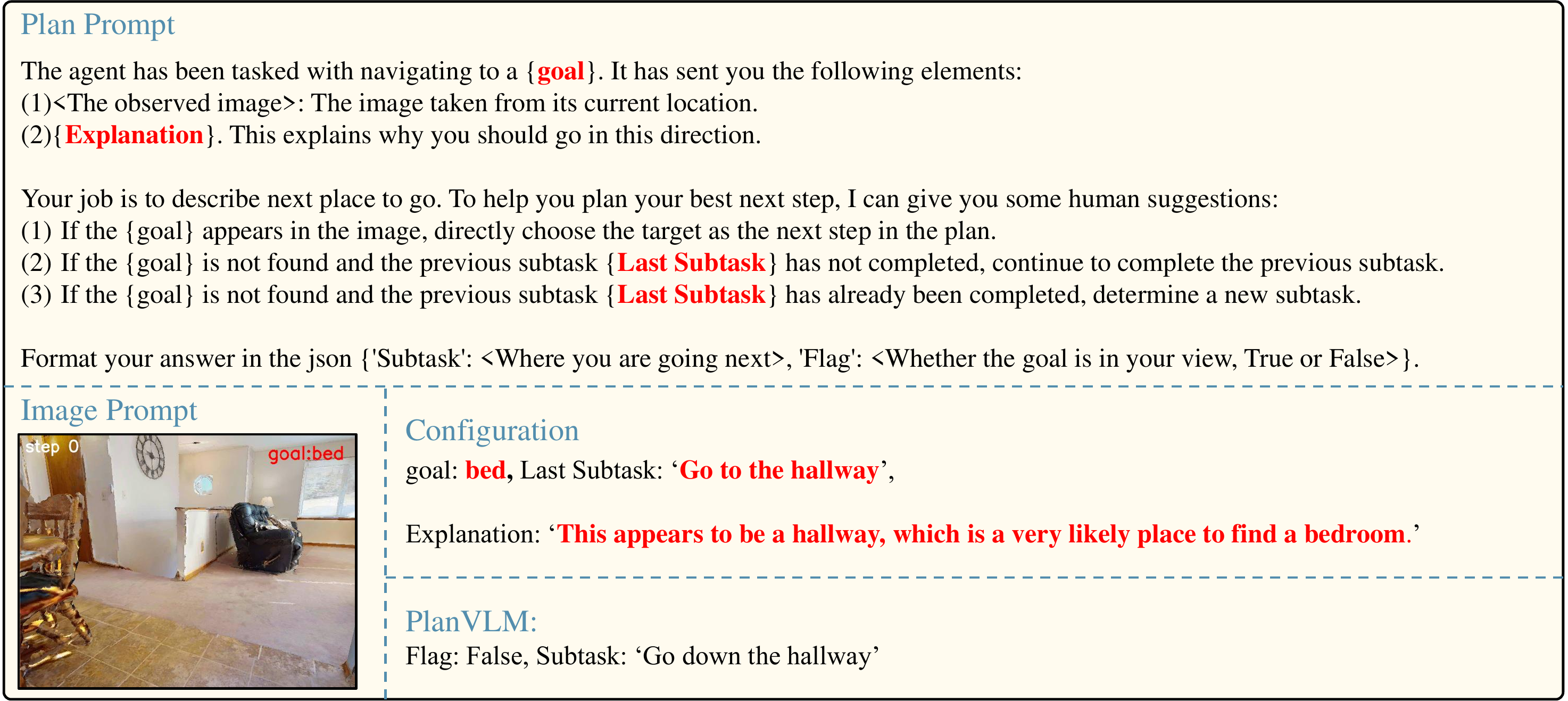}
    \vspace{-3ex}
   \caption{{\bf Plan the Route.} Text prompt is configured by the previous step's subtask, the explanation for selecting the highest-scoring image, and the goal. Using the image with the highest curiosity value as the image prompt, the VLM is invoked to plan the agent's new subtask and detect the goal.}
   \label{fig:Plan}
   \vspace{-13pt}
\end{figure}

Planning a route to an unseen goal in a completely unknown environment is a challenging task because it is hard for the model to get dense rewards. Decomposing the final goal and identifying an intermediate subtask at each step is helpful. For instance, when searching for a bed, an efficient approach would be first finding the corridor leading to the bedroom, then reaching the bedroom door, and finally approaching the bed. We adopt a subtask decomposition strategy to obtain more feedback from the environment. After the updated curiosity value map is stored in the memory, it is projected back to the current navigable areas, and the average curiosity value for each direction is calculated to obtain the final curiosity value score $\overline{Score}_t$. We then select the navigable region corresponding to the direction $\overline{\alpha}$ with the highest score. After selecting a direction $\overline{\alpha}$ based on the curiosity value, we perform more specific planning on this image as shown in~\Cref{fig:Plan}. Specifically, we input the selected image $I_t(\overline{\alpha})$ and the previous subtask $SubTask_{t-1}$ into PlanVLM and it outputs a new subtask $SubTask_t$ and the goal flag $GoalFlag_t$ ($c_t$ in~\Cref{fig:method}):
\begin{equation}
    SubTask_t, GoalFlag_t = PlanVLM(I_t(\overline{\alpha}), SubTask_{t-1})
\end{equation}

\subsection{Two-stage Action Proposer} \label{sec:tap}
\begin{figure}[t]
  \centering
         \includegraphics[width=0.45\textwidth]{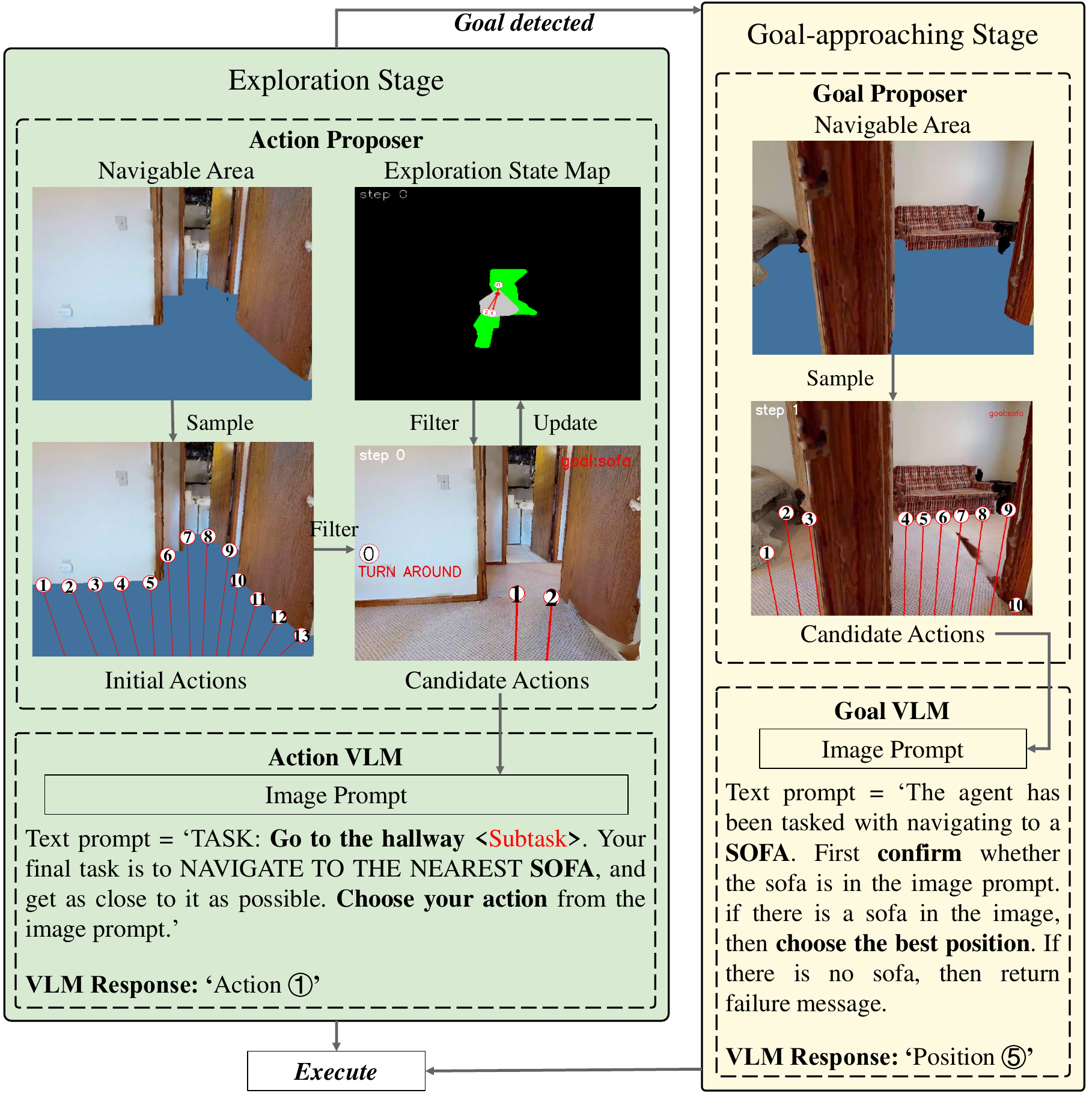}
    \vspace{1ex}
  \caption{{\bf Reason the Action.} In the exploration stage, the agent uses the action proposer to filter sampled actions. ActionVLM(obtained by configuring ReasonVLM) selects the most appropriate action for execution from the image with a labeled candidate action sequence, continuing until the target is found and the agent shifts to the next stage. The next stage is the goal-approaching stage. The agent uses the Goal Proposer to densely sample actions from the image. The GoalVLM(also obtained by configuring ReasonVLM) then selects the action that best represents the goal location.
}
  \label{fig:Reason}
  \vspace{-7pt}
\end{figure}

We employ an action proposer~\cite{vlmnav} to prepare action choices for ReasonVLM. The polar coordinate action space is sampled from the navigable area. For the selected navigable area in the image $I_t(\overline{\alpha})$, we first sample $K$ vectors at regular angular intervals $\Delta \theta$ within the navigable area and map them to the agent's coordinate system to generate an initial action sequence $A^{init}_t=\{(r_{t, i}, \theta_{t, i})\}_{i=1}^K$. Then, actions falling within explored regions are filtered out based on the exploration state map, and the action sequence is further refined by limiting the movement distance and angular intervals. This results in a final candidate action sequence consisting of $K'$ actions. The candidate action sequence $A^{cand}_t=\{(r_{t, j}, \theta_{t, j})\}_{j=1}^{K'}$ in the agent's coordinate system is then mapped back to the image and annotated to obtain $I^{ann}_t$, as shown in~\Cref{fig:Reason}. 

In order to make the agent more purposeful when choosing actions and reduce confusion of VLMs in accurately estimating the distance of objects in images, we divide the entire action-decision process into two stages as in~\Cref{fig:Reason}. The first stage is the exploration stage, where the task is to explore the regions most likely to contain the goal, ultimately discovering and accurately locating its position. The second stage is the goal-approaching stage, where the task is to move as close as possible to the goal location and eventually stop at the goal.
\subsubsection{Exploration Stage}\label{sec:exploratory stage}
Given $SubTask_t$ and $I^{ann}_t$, to satisfy the requirements of the subtask $SubTask_t$, the VLM selects the most appropriate action $\overline{a}_t$ from the action sequence $A^{cand}_t$ in $I^{ann}_t$ for execution:
\begin{equation}
    \overline{a}_t=ReasonVLM(SubTask_t, Goal_t, I^{ann}_t)
\end{equation}
The final action satisfies $\overline{a}_t=(\overline{r}_t, \overline{\theta}_t)$.
\subsubsection{Goal-approaching Stage}
Due to the limitations of the existing VLMs' capability, we do not rely on the VLM to estimate the stopping condition directly from the observed image. Instead, we employ a strategy similar to the action proposer to determine the precise location of the goal when the goal appears in the current observation to make the stopping condition more reliable. The length constraint on the polar coordinate vectors is removed, and sampling in the navigable regions is made denser to ensure the presence of vectors that lead to the goal's ground location at the end. This strategy allows for accurate localization of the target. The agent's stopping condition is set as follows:
\begin{equation}
    StopFlag=
    \begin{cases}
        True, &if\,DistanceToGoal < d_{thres} \\
        False, &else.
    \end{cases}
\end{equation}
where $DistanceToGoal$ is the Euclidean distance between the current position and the goal position.

%% file: seg/4_experiments.tex
\section{Experiments} \label{sec:experiments}
\subsection{Datasets and Evaluation Metrics}

{\bf Datasets} The HM3D v0.1~\cite{ramakrishnan2021hm3d} is used in the Habitat 2022 ObjectNav challenge, providing 2000 validation episodes on 20 validation environments with 6 goal object categories. The HM3D v0.2~\cite{ramakrishnan2021hm3d} is the new version of HM3D with higher quality. It improves geometry and semantic labels with typo fixes, painting error corrections, and has 1000 validation episodes. MP3D~\cite{Matterport3D} contains 11 high-fidelity scenes and 2195 episodes for validation, with 21 categories of object goals.

\input{table/main_results}

{\bf Metrics} We adopt Success Rate (SR) and Success Rate Weighted by Inverse Path Length (SPL) as the evaluation metrics. SR represents the percentage of episodes that are completed. SPL quantifies the agent's navigation efficiency by calculating the inverse ratio of the actual path length traversed to the optimal path length weighted by success rate.

\subsection{Implementation Details}
We set 40 as the agent’s maximal navigation steps. The agent adopts a cylindrical body of radius 0.18m and height 0.88m. We equip the agent with an egocentric RGB-D camera with resolution $640\times 480$ and an HFoV of $79^{\circ}$. The camera is tilted down with a pitch of $14^{\circ}$, which helps determine navigable areas. $d_{thres}$ is set to 1.0, which means if the agent stops when the distance to the goal is less than 0.1m, the episode is considered successful. We mainly use Gemini VLM for our experiments, given its low cost and high effectiveness.

\subsection{Comparison with SOTA Methods}
In this section, we compare our WMNav method with the representative methods for object navigation on the MP3D~\cite{Matterport3D} and HM3D~\cite{ramakrishnan2021hm3d} benchmarks. As shown in \Cref{tab:main results}, our method outperforms all the state-of-the-art zero-shot methods (+3.2\% SR and +3.2\% SPL on HM3D, +13.5\% SR and +1.1\% SPL on MP3D). Compared to all methods, including supervised methods, our approach also achieves the optimal SR on MP3D and the best SPL on HM3D, demonstrating the effectiveness of our method.

Most existing zero-shot methods utilize carefully constructed semantic maps to obtain spatial layout information and utilize Foundation Models (LLMs or VLMs) for commonsense reasoning. Both ESC~\cite{zhou2023esc} and L3MVN\cite{yu2023l3mvn} adopt the frontier map exploration strategy and leverage an LLM to select appropriate frontier points. ESC~\cite{zhou2023esc} converts the image into text scene information for LLM to reason. But textual information cannot accurately describe the spatial relationships in the scene, and it is difficult for LLM to make good spatial decisions. OpenFMNav~\cite{kuang2024openfmnav} uses detectors (Grounding DINO\cite{liu2023grounding} and SAM\cite{kirillov2023segment}) and VLMs to process the image to obtain an informative semantic map. Yet, building a fine map with the help of detectors is too complicated and does not fully stimulate the VLM's ability to understand the scene. TopV-Nav~\cite{zhong2024topv} directly takes the top-view map as the VLM's input to utilize the complete spatial information. However, since VLM is trained on egocentric image data, it does not take advantage of VLM's powerful egocentric reasoning ability. Similar to the frontier map, our simple and online maintained Curiosity Value Map, without prior information from other detectors, makes full use of the scene semantic understanding and predicting ability of VLM to implicitly encode the spatial layout of the scene and predict the results, thus enhancing the navigation efficiency in the unknown environment. 

What's more, methods based on LLM or VLM suffer from hallucination. Therefore, we decompose the task into multiple subtasks and send the previous subtask to VLMs as a cost to reduce misleading hallucinations without fine-tuning the VLM. Our two-stage action proposer strategy avoids reliance on local policy and learning-based policy while utilizing the powerful spatial reasoning ability of VLM. So, the agent only needs a VLM base to complete all the processes without any policy modules to train.
 \input{table/ablation_modules}
\input{table/ablation_vlm}
\subsection{Ablation Study}

\textbf{Effect of Different Modules.}  To manifest the contribution of each module, we compare three ablation models on HM3D V0.2 as it is more representative. Removing TAP implies using only the Action Proposer without computing the goal location after the goal is detected and relying on a stoppingVLM to directly determine the stopping condition according to the observation. As indicated in \Cref{tab:ablation modules}, a and b, a and d, e and f respectively show the effectiveness of modules SubTask Decomposition, Curiosity Value Map, and Two-stage Action Proposer for improving navigation performance.

\textbf{Effect of different VLMs.} We further evaluate the abilities of different VLMs in navigation as shown in \Cref{tab:ablation VLM}, including open-source models and proprietary models. For each row of the table, all the VLMs used in the framework are guaranteed to be of the same model. The comparison between the 3B and 7B models of Qwen2.5-VL~\cite{qwen} shows that an increase in the model's scale can effectively enhance the agent's capabilities. Comparison between Qwen2.5-VL and the Gemini series models reveals that current open-source multimodal models still lag behind advanced proprietary models in terms of ZSON. Gemini 1.5 Pro demonstrates superior capability in the task. Note that when using the smaller Gemini 1.5 Flash VLM, our approach still achieves competitive performance compared to other methods on HM3D, which means our framework is practical on its own rather than relying solely on the capabilities of VLM. What's more, With the evolution of open-source and proprietary VLMs, the ability of each module of our model can be significantly enhanced, and as a whole, it has the potential to show better performance.

\textbf{Effect of different memory strategies.} 
As shown in the \Cref{tab:ablation modules}, we also explore the influence of different memory strategies. No Memory represents that no map memory is used. The Text-Image Memory strategy first uses the VLM to generate textual descriptions of the observation. It constructs a top-down trajectory map and then inputs it as a prompt to the VLM for planning. Curiosity Value Map is our method, which uses a curiosity value map to function as memory. These three strategies all employ SD but do not employ TAP (corresponding to b, c, and e, respectively). The text-image memory strategy performs even worse than the no-map memory strategy. This is because directly feeding the VLM with a text-image combination can easily induce hallucinations, leading to erroneous memory information. Our method shows improvements in both SR and SPL metrics, as the quantitative construction of the Curiosity Value Map forces the VLM to produce output as rigorously as possible and guarantee the accuracy of each call, ensuring reliable memory.

%% file: table/main_results.tex
\begin{table}[t]\small
\small
\caption{\textbf{Zero-shot object navigation results} on HM3D v0.1~\cite{ramakrishnan2021hm3d} and MP3D~\cite{Matterport3D} benchmarks. TF refers to training-free, and ZS refers to zero-shot.}
\label{tab:main results}
\centering
\scriptsize{
\resizebox{\linewidth}{!}{
\setlength\tabcolsep{2.pt}
\renewcommand\arraystretch{1.1}
\begin{tabular}{r||ccIccIcc}
    \hline\thickhline
    \rowcolor{lightgray} &   &  &
    \multicolumn{2}{cI}{\textbf{HM3D}} & \multicolumn{2}{c}{\textbf{MP3D}} \\
    \rowcolor{lightgray} \multirow{-2}*{\textbf{Model}}& \multirow{-2}*{\textbf{TF}}& \multirow{-2}*{\textbf{ZS}}& SR(\%)$\uparrow$ & SPL(\%)$\uparrow$ & SR(\%)$\uparrow$ & SPL(\%)$\uparrow$ \\
            \hline\hline
            Habitat-Web~\cite{Habitat-Web} & \ding{55}  & \ding{55} & 41.5 & 16.0 & 31.6 & 8.5 \\
            OVRL~\cite{ovrl} & \ding{55}  & \ding{55} & - & - & 28.6 & 7.4 \\
            OVRL-V2~\cite{ovrl2} & \ding{55}  & \ding{55} & 64.7 & 28.1 & - & - \\
            \hline\hline
            ZSON~\cite{majumdar2022zson} & \ding{55} & \checkmark & 25.5 & 12.6 & 15.3 & 4.8 \\
            PSL~\cite{sun2024prioritized} & \ding{55} & \checkmark & 42.4 & 19.2 & 18.9 & 6.4 \\
            PixNav~\cite{pixnav} & \ding{55} & \checkmark & 37.9 & 20.5 & - & - \\
            SGM~\cite{sgm} & \ding{55} & \checkmark & 60.2 & 30.8 & 37.7 & 14.7 \\   
            VLFM~\cite{vlfm} & \ding{55} & \checkmark & 52.5 & 30.4 & 36.4 & 17.5 \\
            \hline\hline
            CoW~\cite{gadre2023cows} & \checkmark & \checkmark & - & - & 9.2 & 4.9 \\
            ESC~\cite{zhou2023esc} & \checkmark & \checkmark & 39.2 & 22.3 & 28.7 & 14.2 \\
            L3MVN~\cite{yu2023l3mvn}& \checkmark & \checkmark & 50.4 & 23.1 & - & - \\
            VoroNav~\cite{wu2024voronav} & \checkmark & \checkmark & 42.0 & 26.0 & - & - \\
            OpenFMNav~\cite{kuang2024openfmnav} & \checkmark & \checkmark & 54.9 & 24.4 & - & - \\
            TopV-Nav~\cite{zhong2024topv} & \checkmark & \checkmark & 45.9 & 28.0 & 31.9 & 16.1 \\
            \textbf{WMNav(Ours)} & \checkmark & \checkmark & {\bf 58.1} & {\bf 31.2} & {\bf 45.4} & {\bf 17.2} \\
\hline\thickhline
\end{tabular}}}
\end{table}

%% file: table/ablation_modules.tex
 


\begin{table}[t]\small
\small
\caption{Ablation study of different modules and memory strategies on HM3D v0.2~\cite{ramakrishnan2021hm3d}. SD refers to the subtask decomposition, TAP refers to the Two-stage Action Proposer strategy, No refers to No Memory, Text-Image refers to Text-Image Memory, and CVM refers to Curiosity Value Map.}
\label{tab:ablation modules}
\centering
\scriptsize{
\resizebox{0.85\linewidth}{!}{
\setlength\tabcolsep{2.pt}
\renewcommand\arraystretch{1.1}
\begin{tabular}{cIccc||cc}
    \hline\thickhline
    \rowcolor{lightgray}   & Memory & SD & TAP & SR(\%)$\uparrow$  & SPL(\%)$\uparrow$  \\
 
    \hline\hline

    a&    No & \ding{55} & \ding{55} & 65.8 & 25.8 \\
    b&    No & \checkmark & \ding{55} & 67.4 & 33.1 \\
    c&    Text-Image & \checkmark & \ding{55} & 62.0 & 29.6 \\
    d&    CVM(Ours) & \ding{55} & \ding{55} & 69.5 & {\bf 34.9} \\
    e&    CVM(Ours) & \checkmark & \ding{55} &  70.1& 34.7  \\
    f&    CVM(Ours) & \checkmark & \checkmark & {\bf 72.2}  &  33.3 \\
\hline\thickhline
\end{tabular}}}
\end{table}

%% file: table/ablation_vlm.tex
\begin{table}[t]\small
\small
\caption{Ablation study of VLMs on HM3D v0.1~\cite{ramakrishnan2021hm3d}.}
\label{tab:ablation VLM}
\centering
\scriptsize{
\resizebox{0.72\linewidth}{!}{
\setlength\tabcolsep{2.pt}
\renewcommand\arraystretch{1.1}
\begin{tabular}{r||cc}
    \hline\thickhline
    \rowcolor{lightgray}    VLM &  SR(\%)$\uparrow$ & SPL(\%)$\uparrow$ \\
    \hline\hline
    Qwen2.5-VL-3B & 29.7 & 15.8\\
    Qwen2.5-VL-7B & 46.1 & 20.7\\
     Gemini 1.5 Flash & 53.5 & 27.0 \\
     Gemini 2.0 Flash & 57.9 & 30.7\\
     Gemini 1.5 Pro &{\bf 58.1}  & {\bf 31.2}  \\
\hline\thickhline
\end{tabular}}}
\end{table}

%% file: seg/5_conclusions.tex
\section{Conclusion}
\label{sec:conclusion}
We have introduced WMNav, which finds a novel direction for object goal navigation in unknown environments by leveraging VLMs in a world model framework and significantly enhancing ZSON performance.
Our method addresses serious inefficiency problems of back-and-forth redundant movement by employing an online Curiosity Value Map to quantitatively predict the likelihood of the target's presence.
The subtask decomposition module provides a denser reward for the prompt-based optimization of the policy module. Moreover, the two-stage action proposer strategy leads to more purposeful navigation and efficient exploration.
By constructing the world model architecture based on VLMs, concise memory map building, and task breakdown, WMNav indicates a new optimization direction for the ZSON task and opens up new pathways for embodied robots to interact with environments.

%% file: seg/8_acknowledgment.tex
\section*{ACKNOWLEDGMENT}
This work was supported by the National Natural Science Foundation of China under Grant 62373356 and the Joint Funds of the National Natural Science Foundation of China under U24B20162.